# ADAPTIVE NEURAL CONTROL FOR MOBILE ROBOTS AUTONOMOUS NAVIGATION

**Monica Dragoicea, Ioan Dumitrache, Nicolae Constantin**

*University Politehnica Bucharest*
*Automatic Control and Systems Engineering Dept.*
*Splaiul Independentei 313, 77206 – Bucharest, Romania*
*E-mail: mdragoicea@ics.pub.ro*

**Abstract**: This paper presents a combined strategy for tracking a non-holonomic mobile robot which works under certain operating conditions for system parameters and disturbances. The strategy includes kinematic steering and velocity dynamics learning of mobile robot system simultaneously. In the learning controller (neural network based controller) the velocity dynamics learning control takes part in tracking of the reference velocity trajectory by learning the inverse function of robot dynamics while the reference velocity control input plays a role in stabilizing the kinematic steering system to the desired reference model of kinematic system even without using the assumption of perfect velocity tracking.

## 1. INTRODUCTION

Nowadays practical, technological requirements imposed the need of advanced control technique design as well as a suitable mathematical support, in order to allow addressing complex problems. Some specific reasons could be highlighted here: increased complexity of the present industrial plants / processes, demanding operation requirements, the design of control systems that should take into account important uncertainties (e.g. insufficient a priori information about the process and its operation environment).

The goal of autonomous mobile robotics is to build physical systems that can move toward a goal without human intervention in previously unexplored environments – that is, in real environments that have not been specifically engineered for the robot (Balch, 1998). An important issue in autonomous navigation is the need to cope with the large amount of uncertainty that is an inherent characteristic of natural environments. Based on these observations we can claim that using intelligent techniques (artificial neural networks, fuzzy systems, genetic algorithms, or synergetic combinations of them) we can overcome some drawbacks of current technology in building complex systems with certain degrees of autonomy when dealing with uncertain information.

Intelligent controllers are one way to achieve the declared objective of autonomy (Li, 1999). An autonomous agent must have learning mechanisms built into its controller. This feature will allow the agent to obtain more freedom to act on its own when facing environmental changes (e.g. geometrical configuration alteration or the movement of other mobile objects). The control system must be designed to ensure highly autonomous operation of the control functions. The controller should deal with unexpected situations, new control tasks, and failures within limits. To achieve this, high decision making techniques for reasoning under uncertainty and taking actions must be used. These techniques, if used by humans, are attributed to intelligent behavior. Hence, one way to achieve autonomy, for some applications, is to use high level decision making techniques, intelligent methods, in the autonomous controller. Different architectures of autonomous control systems for mobile robots taking into consideration intelligent techniques like fuzzy logic, neural networks, genetic algorithms were proposed by now.

## 2. PATH TRACKING CONTROL

The aim of this paper is to solve a path tracking control problem for a car-like robot moving on a flat ground. This problem was presented in detail in (Dragoicea, 2000) and will be briefly introduced here. Previous results were presented in (Dumitrache, 1999).

The desired trajectory of the mobile platform on the ground is defined with respect to a cartesian reference coordinate system and it is specified by the reference kinematic equations:

$$\dot{x}_r = V_{AV\_r} \cos\theta_r$$
$$\dot{y}_r = V_{AV\_r} \sin\theta_r$$
$$\dot{\theta}_r = \omega_r$$
$$q_r = [x_r \ y_r \ \theta_r]^T$$
$$\underline{\eta}_r = [V_{AV\_r} \ \omega_r], \quad \underline{\eta}_r > 0, \forall t$$

The navigation problem for a non-holonomic system impose the determination of the control vector $\underline{\eta}_c(t)$ for the orientation and movement of the mobile platform (see figure 1). Figure 1 presents the hierarchical control structure for path tracking control.

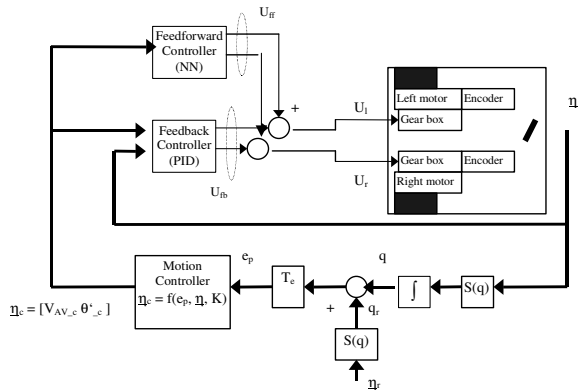

Figure 1 Two level hierarchical control structure

In this respect, one aims to determine a control law $\underline{\eta}_c(t) = f_c(e_p, \underline{\eta}_r, K)$, so that $\lim_{t\to\infty} (q_r - q) = 0$, where $e_p$ is the positioning error, $\underline{\eta}_r$ is the velocity reference vector and K is the vector of the controller parameters. Based on $\underline{\eta}_c(t)$, the torques vector $\tau(t)$ will be determined so that $\underline{\eta}(t) \to \underline{\eta}_c(t)$ when $t \to \infty$. It is the case when the mobile vehicle has to arrive to the goal position in a specified time interval.

This paper presents a combined strategy for tracking a non-holonomic mobile robot which works under certain operating conditions for system parameters and disturbances (see figure 1). The strategy includes kinematic steering and velocity dynamics learning of mobile robot system simultaneously. In the learning controller (neural network based controller) the velocity dynamics learning control takes part in tracking of the reference velocity trajectory by learning inverse function of robot dynamics while the reference velocity control input plays a role in stabilizing the kinematic steering system to the desired reference model of kinematic system even without using the assumption of perfect velocity tracking.

## 3. TRAJECTORY – TRACKING CONTROLLER

The trajectory - tracking controller should produce a pair of control inputs $\underline{\eta}_c = [V_{AV}{}^c \ \dot{\theta}^c]$ (see figure 1) and guarantee that the trajectory of the car-like robot driven by $\underline{\eta}_c$ converges to the desired trajectory, i.e.

$$\lim_{t\to\infty}(q_r - q) = 0 \tag{1}$$

where $q_r = [x_r \ y_r \ \theta_r]$ is the reference posture of the robot, $q = [x \ y \ \theta]$ is the coordinate vector that determines the current position of the robot in plane.

We can formulate the trajectory - tracking controller as follows. The corresponding coordinate transformation that defines the tracking error system (Kanayama, 1990) is given by:

$$e_p = T_e(q_r - q)$$
$$\begin{bmatrix} e_x \\ e_y \\ e_z \end{bmatrix} = \begin{bmatrix} \cos\theta & \sin\theta & 0 \\ -\sin\theta & \cos\theta & 0 \\ 0 & 0 & 1 \end{bmatrix} \begin{bmatrix} x_r - x \\ y_r - y \\ z_r - z \end{bmatrix} \tag{2}$$

Taking the time derivative of (2) and using the kinematic constraints:

$$\dot{x} = V_{AV} \cos\theta$$
$$\dot{y} = V_{AV} \sin\theta \tag{3}$$
$$\dot{\theta} = \omega_r$$

the following differential equation of the tracking error system can be obtained:

$$\dot{e}_p = f(e_p) + g(e_p)\underline{\eta}_c \tag{4}$$

where

$$e_p = [e_x \ e_y \ e_\theta]^T, \quad f(e_p) = \begin{bmatrix} V_{AV\_r} \cos e_\theta \\ V_{AV\_R} \sin e_\theta \\ \omega_r \end{bmatrix}$$

$$g(e_p) = \begin{bmatrix} -1 & e_y \\ 0 & -e_x \\ 0 & -1 \end{bmatrix}, \quad \underline{\eta}_c = \begin{bmatrix} V_{AV}^c \\ \omega^c \end{bmatrix} \tag{5}$$

Therefore, the aim of the trajectory tracking control is to stabilize the system of (4) to the origin. We propose the following controller that can stabilize the nonlinear system. For the above tracking error system (0,0,0) is globally asymptotically stable equilibrium point if the control inputs are:

$$\begin{cases} V_{AV}^c = k_1 e_x + V_{AV\_r} \cos e_\theta \\ \omega^c = \omega_r + \dfrac{V_{AV\_r}}{2} k_2 (e_y + k_3 e_\theta) + \dfrac{V_{AV\_r}}{2k_3} \sin e_\theta \end{cases} \quad (6)$$

where $k_1$, $k_2$, $k_3$ are the parameters of the tracking controller (positive constants) and $V_{AV\_r} > 0$.

Indeed, considering the following Lyapunov function:

$$V(e_p) = \frac{1}{2} e_x^2 + \frac{1}{2}(e_y + k_3 e_\theta)^2 + \frac{1}{k_2}(1 - \cos e_\theta) \quad (7)$$

the time derivative of $V(e_p)$ is

$$\dot{V}(e_p) = -k_1 e_x^2 - 0.5 V_{AV\_r} k_2 (e_y + k_3 e_\theta)^2 - 0.5 V_{AV\_r} k_2 k_3 (e_y + k_3 e_\theta)^2 - 0.5 V_{AV\_r} \frac{1}{k_3^2} \sin^2 e_\theta \quad (8)$$

Clearly, $V \geq 0$. If $e_p \equiv 0$, $V = 0$. If $e_p \neq 0$, $V > 0$. Furthermore, by equation (8), $\dot{V} \leq 0$. So, $V(e_p)$ is a Lyapunov function.

Therefore, the control law presented in equation (6) is a solution for the problem of trajectory - tracking controller implementation. Using different controller parameters $K = [k_1, k_2, k_3]$, we can obtain different convergence performances.

## 4. NEURAL NETWORKS BASED CONTROL STRATEGY

The present paper formulates a control strategy for mobile robot control based on artificial neural networks (ANN) at the inner level of velocity control. The approach considered in the present paper combines the stabilization properties of a feedback conventional PID controller with the learning capabilities of a neural network based feed-forward controller.

Conceptually, most of the approaches that accounts for neural networks based control structures make use of the "inverse" model of the process as a controller. In this article the feed-forward controller is implemented as a neural network controller that represents the inverse model of the process.

A feed-forward adaptive control strategy based on neural networks is used in order to implement the velocity controller. Figure 1 depicts the two components of the velocity controller:

- a feedback component (PID control of linear and angular velocities, $V_{AV}$ and $\dot{\theta}$), based on a simplified model of the process (dc motors and mobile platform) (Mazo, 1995); it will stabilize the process and will reject the perturbations.
- a feed-forward component, with learning abilities, will allow a fast trajectory tracking. This component is important in improving the system performances by learning on-line information about the process, through direct interaction. Actually, this component will learn on-line the inverse dynamics of the process.

In this way, it is possible to implement an adaptive control strategy at the execution level, based on neural networks. The neural network (feed-forward controller) generates a command signal $U_{ff}$ that will adjust the signal generated by the feedback controller, $U_{fb}$, in order to minimize the velocity reference error, while compensating the modeling uncertainties. The original contribution of this work consists in obtaining the command vector $\underline{u} = [U_l\ U_r]$, using the reference velocity vector $\underline{\eta}_c$ from the motion controller (Dumitrache, 2001). A new possibility of choosing the motion controller was also presented. The motion controller was implemented following the procedure presented in section 3, following an idea presented in (Kanayama, 1990). The mobile robot can be considered to be a two inputs - (the torques for the dc. motors) - two outputs (the linear velocity along the main axis, $V_{AV}$, and the rotation velocity, $\dot{\theta}$) system (Dragoicea, 2001).

## 5. RESULTS

The obtained results were divided in two categories: one that explains the way the feed-forward control strategy can improve the tracking of the references velocities (linear and angular velocities, as well as right and left wheels velocities) and a second one that proves effective results in tracking a desired trajectory based on the method presented in section 3.

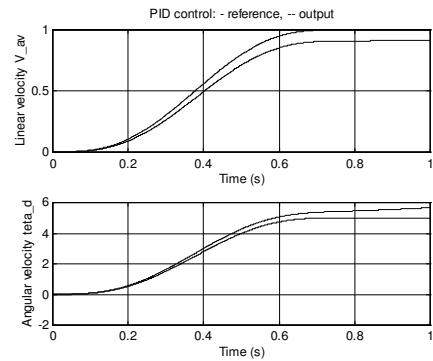

Figure 2 Velocity control – PID control

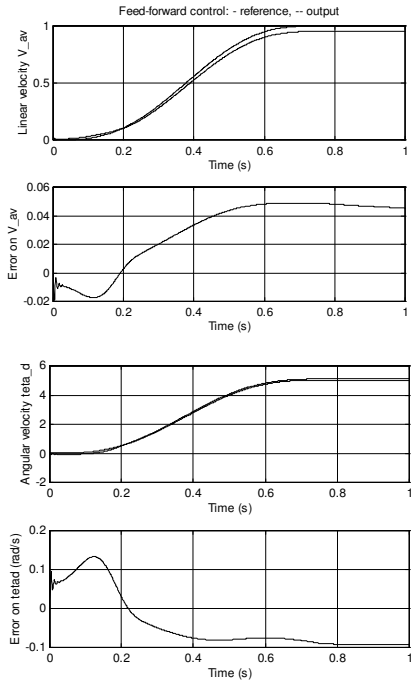

Figure 3 Velocity control – feedforward control

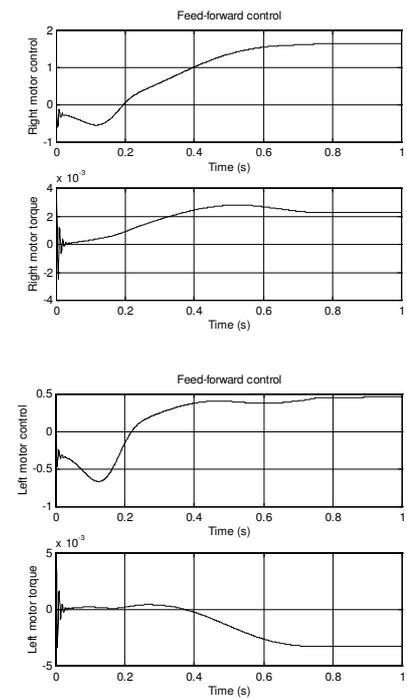

Figure 4 Feed-forward control – dc motors control

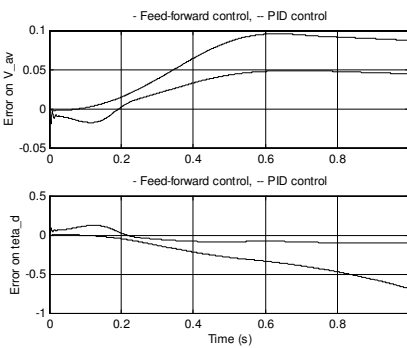

Figure 5 Errors on velocities

The following three figures demonstrates that the control law for trajectory control (equation (6)) is effective.

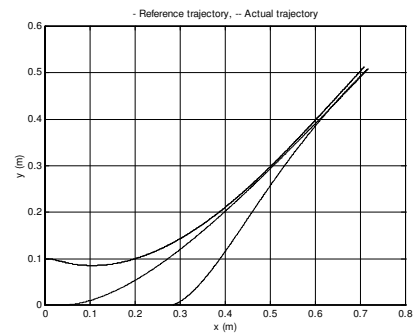

Figure 6 Trajectory – tracking;
a)  $q_o = [0.3\ 0\ –5^o]$,  K = [2.3 0.3 3.8]
b)  $q_o = [0\ 0.1\ –10^o]$, K = [5 5 0.1]

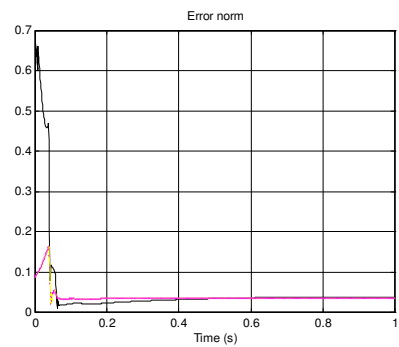

Figure 7 Error norm, $\underline{e}_c(t) = \underline{\eta}_c(t) - \underline{\eta}(t)$

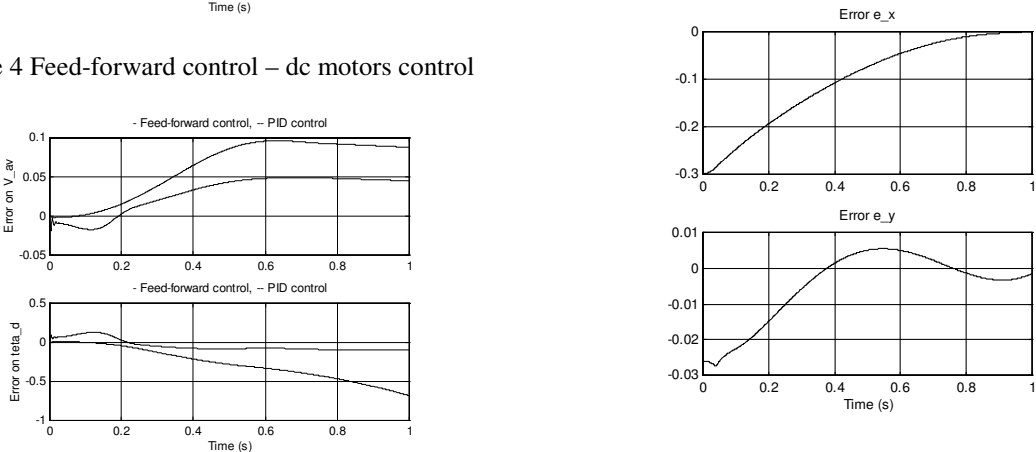

Figure 8 Positioning errors, $\underline{e}_p$

As can be easily observed in figure 7, the norm of the error is bounded and very small. That means that the implemented control strategy allows to obtain a stable operation, all signals being bounded (see also figure 8). Performance requirements are also fulfilled as can be seen in figure 6.

The analysis of the stability properties of the whole adaptive control structure for trajectory control with neural networks velocity control was carried out for different operating conditions, initial conditions $q_o$ and controller parameters K = [$k_1$ $k_2$ $k_3$]. The results are presented selectively for certain situations, but the general evolution of the signal in the system is always the same for each of the considered cases.

Some more observation can be further highlighted. The tracking performances are sensible dependent on choosing the parameters of the tracking controller, K; this choice should be carefully correlated with the selection of the initial condition for movement, $q_o$. In this work the selection of the trajectory controller parameters was made by trial-and-error methods. It was observed that a larger value of $k_1$ leads to a faster convergence and to the reduction of the positioning error $e_x$.

## 6. CONCLUDING REMARKS

This paper presents the implementation of a control structure for trajectory control of a mobile platform based on intelligent control techniques. The classical control structure for velocity control was extended using artificial neural networks in order to take into consideration the modeling uncertainties of the controlled process. An adaptive control system for a mobile robot was implemented that is able to compensate for unmodelled dynamics and parametric uncertainties.

The advantage of this control strategy is that the redefinition of the input control $\eta c(t)$ allows to obtain different stable behaviors of the mobile platform, e.g. path-tracking control. The adaptive control strategy at the low level control will guarantee the performances when process parameters change, because the dynamic controller (neural network controller) together with the trajectory controller will act to compensate the lack of a priori information about the process.

In conclusion, while the inferior level (velocity control) compensates for the modeling uncertainties and perturbations, in the outer level (tracking control) different techniques for planning and movement control can be implemented. In this way, the control strategy can be easily extended. Therefore, the mobile robot can obtain a desired degree of autonomy by the integration of intelligent control techniques.